\documentclass[10pt,twocolumn,letterpaper]{article}

\usepackage{wacv}
\usepackage{times}
\usepackage{epsfig}
\usepackage{graphicx}
\usepackage{amsmath}
\usepackage{amssymb}
\usepackage{placeins}
\usepackage{soul}

\usepackage[pagebackref=true,breaklinks=true,letterpaper=true,colorlinks,bookmarks=false]{hyperref}

\wacvfinalcopy 

\def\wacvPaperID{128} 

\ifwacvfinal\fi
\begin{document}

\title{Unsupervised Feature Learning of Human Actions as Trajectories in Pose Embedding Manifold}

\author{Jogendra Nath Kundu\thanks{equal contribution - listed alphabetically by first names} \qquad Maharshi Gor\footnotemark[1] \qquad Phani Krishna Uppala \qquad R. Venkatesh Babu\\
Video Analytics Lab, CDS, Indian Institute of Science, Bangalore, India\\
{\tt\small jogendrak@iisc.ac.in, \{maharshigor18, krishnaphaniiitg\}@gmail.com, venky@iisc.ac.in}}





\maketitle
\ifwacvfinal\thispagestyle{empty}\fi

\begin{abstract}
   An unsupervised human action modeling framework can provide useful pose-sequence representation, which can be utilized in a variety of pose analysis applications. 
   In this work we propose a novel temporal pose-sequence modeling framework, which can embed the dynamics of 3D human-skeleton joints to a continuous latent space in an efficient manner. In contrast to end-to-end framework explored by previous works, we disentangle the task of individual pose representation learning from the task of learning actions as a trajectory in pose embedding space. In order to realize a continuous pose embedding manifold with improved reconstructions, we propose an unsupervised, manifold learning procedure named Encoder GAN, (or \textit{EnGAN}). Further we use the pose embeddings generated by \textit{EnGAN} to model human actions using a bidirectional RNN auto-encoder architecture, \textit{PoseRNN}. We introduce first-order gradient loss to explicitly enforce temporal regularity in the predicted motion sequence. A hierarchical feature fusion technique is also investigated for simultaneous modeling of local skeleton joints along with global pose variations. We demonstrate state-of-the-art transfer-ability of the learned representation 
   against other supervisedly and unsupervisedly learned motion embeddings for the task of fine-grained action recognition on SBU interaction dataset.
Further, we show the qualitative strengths of the proposed framework by visualizing skeleton pose reconstructions and interpolations in pose-embedding space, and low dimensional principal component projections of the reconstructed pose trajectories.
\end{abstract}

\section{Introduction}

\label{sec:intro}
Supervised deep learning methods have exhibited promising results for the task of action recognition from 3D skeletal pose ~\cite{zhang2017view,liu2016spatio,hu2017temporal}. However performance of these methods highly relies on availability of large number of diverse data samples with annotated action labels. From the perspective of unsupervised feature learning, action-label agnostic feature representation can generalize to various novel classes in contrast to the acquired feature representation from a supervised counterpart.  Moreover, a generalized representation of temporal pose dynamics can be used as an initial step to facilitate various different tasks like analytics
of gymnastics, dance motion data, sports bio-mechanics, 3D film production, motion transfer to humanoid robots etc. Under the umbrella of unsupervised learning methods, a generative approach to model the temporal dynamics of human pose can easily be extended not only for action recognition but also for temporal pose generation and future prediction tasks. Hence, there is a need for an efficient unsupervised pose modeling approach which can serve various different tasks with an improved transferability. In contrast to an end-to-end framework~\cite{taku_motion_synthesis,butepage2017deep} for representation and synthesis of pose dynamics, we propose a novel pose-sequence modeling strategy. Previous works ~\cite{martinez2017human,li2017auto} train a single end to end model for handling two different complex tasks of predicting plausible human pose, along with modeling the temporal dynamics and hence, are not scalable. In contrast to such black-box approaches, we plan to separately model a distribution of plausible human poses and further use that to model the sequential dynamics of an action.

We propose a novel generative adversarial network (GAN) with an encoder setup designed specifically to have a good hold on the latent representation $z$, which can produce a given 3D pose, $x$. In a usual GAN setup, one optimizes over the latent representation $z^*$ to generate a reconstructed pose $x^\prime = G(z^*)$ such that $|x - x^\prime|$ is minimized \cite{yeh2016semantic}. Note that this is an iterative optimization process and hence time consuming. To avoid such iterative process in later steps we simultaneously train an encoder $Pose^{enc}$ such that, $z = Pose^{enc}(x)$ can be obtained in a single inference. In the proposed Encoder-GAN (\textit{EnGAN}) setup we simultaneously learn the generator, $Pose^{dec}(z)$ along with the encoder $Pose^{enc}(x)$. This leads us to the question - why not use a simple auto encoder instead of \textit{EnGAN}? One of the major benefit of employing a GAN framework is that, it can learn a continuous latent embedding subspace in contrast to the latent space learned by a simple auto-encoder setup. Hence, \textit{EnGAN} enables us to randomly sample plausible human pose interpolations in the continuous embedding space between two diverse pose representations. This also facilitates better modeling of pose dynamics, represented as a continuous trajectory in the learned embedding space. Detailed training procedure of \textit{EnGAN} is discussed in Section \ref{sec:engan}. Parenthetically, the proposed pose embedding model can also be used in applications related to 3D pose estimation to deliver plausible 3D pose from various types of 2D projection information.

In this paper we model skeleton sequence dynamics, representing a particular action as a trajectory in the pose embedding space. For long temporal sequences a single Recurrent Neural Network (RNN) fails to efficiently model both short-term and long-term trajectory variations. However, in the available action recognition datasets, the cues specific to a particular action might have performed for a short time span. Thus, we employ a stacked two layer bidirectional RNN to effectively model diverse pose dynamics by designing a RNN auto-encoder architecture, \textit{PoseRNN}. Couple of recent approaches~\cite{li2017auto,butepage2017deep,taku_motion_synthesis} have also explored such RNN auto-encoder setup but in a fully end-to-end fashion. These approaches provide raw joint coordinates directly to the RNN encoder and consequently the decoder is trained to reconstruct back the raw skeleton joint coordinates. In contrast to such approaches we absolutely disentangle the learning of pose embedding from the learning of temporal pose dynamic in a much efficient manner. Note that, in the proposed \textit{PoseRNN} we feed the sequence of pose embedding feature and following this the decoder is trained to deliver the pose embedding feature instead of the raw joint coordinates. This disentanglement of tasks allows us to train the pose modeling network in a complete unsupervised setup, over a large amount of unannotated human 3D skeleton data. We also explore a hierarchical feature fusion technique to fuse local joint level features and individual limb representations along with the global pose embedding. This helps to effectively address action samples focusing on local joint dynamics such as; playing with a tablet, typing on a keyboard, writing, etc. We also incorporate a direct loss on the predicted skeleton joint with an additional first order gradient based loss to explicitly encourage temporal regularity. Finally we demonstrate effectiveness of the learned trajectory embedding for the task of action recognition in multiple datasets with minimal supervision on annotated samples.

\noindent
In summary, our main contributions are as follows; 
\begin{itemize}
\vspace{-1mm}
\item  A novel generative architecture for human pose data (\textit{EnGAN}), which can effectively learn a continuous latent space simultaneously with an encoder setup facilitating one-shot inference. 

\item A novel method of \textit{hierarchical feature fusion} with loss on the end task of skeleton joint prediction followed by incorporation of \textit{first order temporal regularity} to improve human motion generation, which can facilitate learning of more general motion features. 

\vspace{-3mm}
\item Clear demonstration of effectiveness of both \textit{EnGAN} and the \textit{EnGAN-PoseRNN} combination against available unsupervised approaches.  We also demonstrate \textit{state-of-the-art transferability} of the learned representation against other supervisedly and unsupervisedly learned motion embeddings for the task of fine-grained action recognition on SBU interaction dataset.
\end{itemize}

\section{Related Work}
\label{sec:related_work}

\subsection{Supervised action recognition}
There is a cluster of previous arts on the use of Recurrent Neural Network (RNN) based models for the task of action recognition from sequence of 3D skeleton joint coordinates. Du \etal ~\cite{du2015hierarchical, du2016representation} proposed an hierarchical end-to-end bidirectional RNN architecture inspired from the kinematic tree representation of limb connections. They use LSTM subnetwork to model 5 different body parts viz. two arms, two legs and one trunk and hierarchically combine limb representations in further layers. For efficient learning of part-based dynamics, Shahroudy \etal ~\cite{shahroudy2016ntu} propose to initially split the memory cell of LSTM into part-based sub-cells followed by late fusion of features for action recognition instead of modeling the full body as a whole. Zhu \etal ~\cite{zhu2016co} propose a co-occurrence learning regularization approach by introducing fully connected layer between LSTM network to encourage learning of general connections without using kinematic tree supervision. Liu \etal ~\cite{liu2016spatio} introduced a new gating technique in LSTM to explicitly handle noise and occlusion in input data. They also extended LSTM architecture to work on the spatio-temporal domain to facilitate efficient modeling of joint dependencies. To further utilize the action specific local cues in an explicit fashion, Song \etal ~\cite{song2017end} adopted a spatio-temporal attention mechanism to selectively focus on discriminative joints in a frame along with an additional attention level on representations from different time steps. Hu \etal ~\cite{hu2017temporal} proposed temporal perceptive network (TPNet), where they designed a temporal convolutional subnetwork embedded between the RNN layers to efficiently model short-term pose dynamics. To explicitly model joint dependencies along with temporal pose dynamics, Wang \etal ~\cite{wang2017modeling} proposed a two-stream RNN architecture with different streams for temporal dynamics and spatial configuration.

\subsection{Human motion synthesis.}
Here we provide an overview of the previous works related to 3D skeleton synthesis or prediction. To learn a manifold of human motion data, Holden \etal ~\cite{holden2015learning} used convolutional autoencoders to learn the prior probability distribution of plausible pose representation. Akhter \etal \cite{akhter2015pose} proposed pose-prior by learning pose-dependent joint angle limits, which can be used to avoid prediction of unambiguous 3D pose. Fragkiadaki \etal ~\cite{fragkiadaki2015recurrent} proposed an Encoder-Recurrent-Decoder (ERD) architecture for jointly learning skeleton embedding along with temporal pose forecasting. Recently Martinez \etal ~\cite{martinez2017human} proposed     a sequence-to-sequence architecture with residual connections to model the short-term motion predictions using sampling based loss. In the proposed unsupervised human motion modeling approach we have followed an entirely new direction. The disentanglement of pose modeling with respect to the sequential nature of pose dynamics not only achieves improved pose generation results but also learns a generalized trajectory embedding space, delivering state-of-the-art transferability for action recognition performance.

\section{Approach}
\label{sec:approach}
In this section we describe the proposed pose manifold learning methodology along with the carefully carried out preprocessing steps to make \textit{EnGAN} invariant to translation, view and scale, both at global as well as at skeleton-joint level. The adopted preprocessing steps not only improves learning of efficient pose manifold but also facilitates efficient training of \textit{PoseRNN} encoder-decoder setup by exploiting the disentangled canonical-pose and global position parameters. In further part of this section  we elaborate the learning procedure of \textit{PoseRNN} followed by utilization of hierarchical feature fusion relevant for fine-grained action recognition. 

\subsection{Canonical pose representation}
\label{sec:preprocessing}
The raw X, Y, Z positions of the skeleton-joints in the world coordinate-system are converted to root relative joint positions, by shifting the origin to the pelvis joint.

On each temporal sequence of these Root Relative joint positions, Savitzky–Golay filter is applied for motion smoothing. This is followed by Kinematic Skeleton Fitting, where each joint is represented in Polar Coordinate system with reference to its parent joint, in the kinematic skeleton tree. Scale of each skeleton sample is normalized to ensure fixed length of a particular limb across all pose samples.

Further, we define a View Invariant, \textit{Skeleton Coordinate System}, whose coordinate axes (X, Y and Z) are represented in terms of the skeleton-joint positions in the \textit{Root Relative Coordinate System}. A sequential rotational transform is applied about X, Y and Z axes, with the angular changes $\alpha$, $\beta$ and $\gamma$ respectively, on the joint positions in the Root Relative Coordinate System to get the corresponding skeleton in View Invariant \textit{Skeleton Coordinate System}.

Now, keeping the location of torso joints fixed, we represent the remaining joints in the coordinate system defined at their respective parent joints, using  Global to Local Coordinate Conversion~\cite{akhter2015pose}. This is done to capture the most relevant joint-level local variations, which are agnostic to the changes in the skeleton at global level. We define this final form as \textit{Canonical Pose Representation}.

The above mentioned carefully selected preprocessing steps helps to model accurate priors of 3D human pose independent of global variations. Our pose embedding model also follows a hierarchical limb based feature fusion to efficiently model correlation between joints and limbs. This facilitates learning of an embedding space with improved generalization avoiding invalid 3D pose samples. 

\begin{figure*}
\begin{center}
	\includegraphics[width=0.8\linewidth]{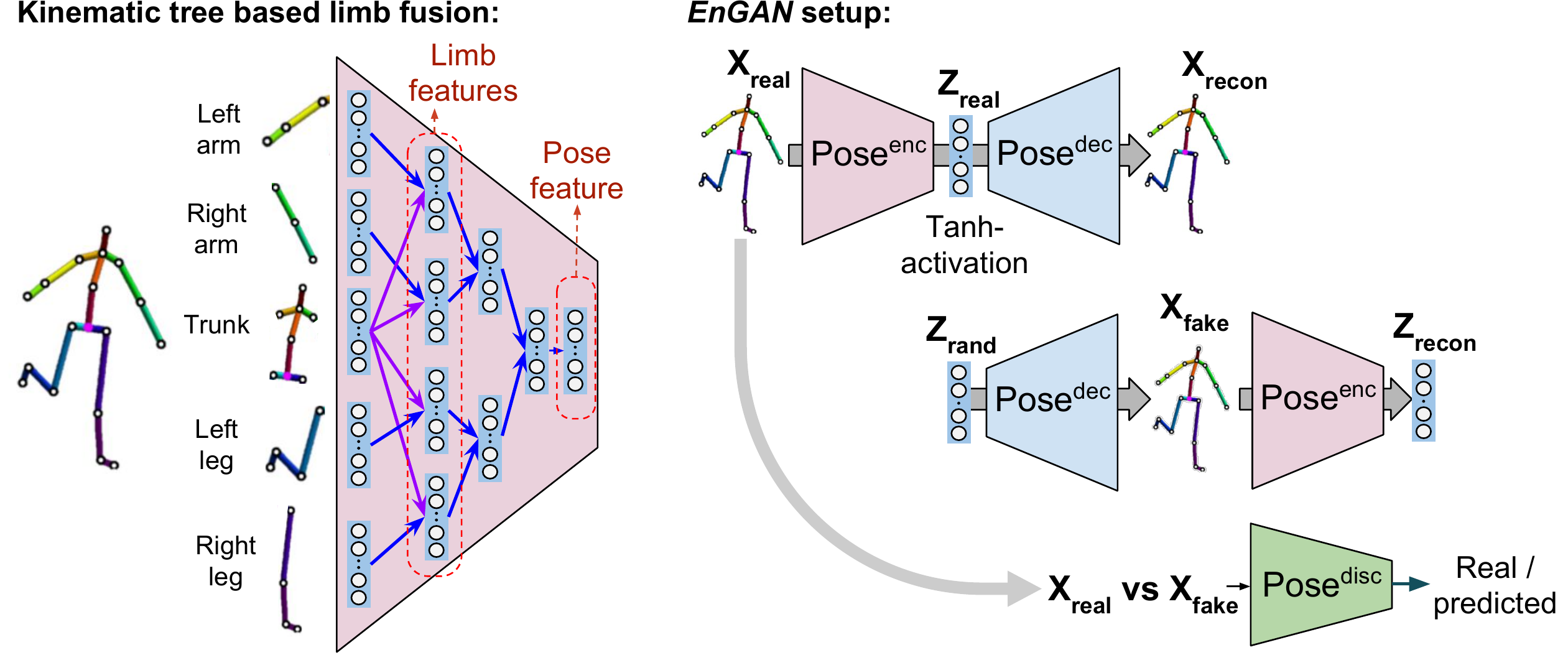}
	\caption{Illustration of EnGAN Pipeline for learning skeleton pose embedding space in a hierarchical way}
 	\label{fig:fig_2}    
\end{center}
\vspace{-3mm}
\end{figure*}

\subsection{Learning pose embedding framework \textbf{\textit{EnGAN}}}
\label{sec:engan}
Consider $x_{real} \in X_{real}$ as a sample of canonical pose representation obtained from the above mentioned transformations on the raw 3D skeleton joint coordinates. Considering $p(x_{real})$ as the distribution of canonical pose representation, we plan to learn a latent representation $z_{real} \in Z_{real}$ such that, $z_{real} = Pose^{enc}(x_{real})$. Here, we constraint distribution of latent representation to be in the domain of $[-1, 1]$ along all the 32 dimensions of latent representation $z$. One of major distinguishing factor of \textit{EnGAN} is that, it efficiently learns the backward projection from $X$ to $Z$ simultaneously with the learning of a continuous latent manifold as seen in case of  Generative Adversarial Networks (GAN). Moreover Variational Autoencoder (VAE) is also not a suitable choice with regard to the specific requirement of an efficient back projection, i.e. $z = Pose^{enc}(x)$. In a VAE setup, efficient generation capability with continuous latent manifold modeling is achieved by careful balancing of both Kullback-Leiber divergence and reconstruction error term in the final objective function. Also, the encoder network of a VAE architecture predicts the parameters of the distribution $p(z|x)$ and hence introduces uncertainty in prediction of exact $z$ which can produce the given $x$. A straightforward autoencoder model lacks in learning a continuous pose manifold and thus does not support an efficient interpolation or traversal in the learned latent space. To alleviate the above mentioned drawbacks of standard approaches, we propose a novel learning protocol to achieve the specific requirement of minimum reconstruction error simultaneously with an efficient manifold modeling.

As shown in right section of Figure \ref{fig:fig_2}, the entire \textit{EnGAN} setup constitutes of 3 different individual networks namely; a) pose encoder $Pose^{enc}$, b) pose decoder $Pose^{dec}$ and c) pose discriminator $Pose^{disc}$. The architectural design is similar across all these three networks, which is inspired from the kinematic tree of limb connections as shown in the left section of Figure \ref{fig:fig_2}. Motivated from the work of Du \etal ~\cite{du2015hierarchical}, we follow a similar hierarchical fusion of limb features for individual pose frames in contrast to their approach of fusing temporal information at different hierarchical levels using bidirectional RNN. The five different body parts viz. two arms, two legs and one trunk are first processed independently and further fused in later hierarchy namely, upper-body and lower-body representations. Towards the end a single representation for the full-body is achieved by fusing both upper and lower body features.

The training procedure of \textit{EnGAN} can be broadly divided into three consecutive steps. Motivated from the CycleGAN~\cite{zhu2017unpaired}  configuration, we first train a cycle-autoencoder without the adversarial discriminator loss. We sample $x_{real}$ from a canonical-pose transformed dataset whereas $z_{rand}$ is sampled randomly from an uniform distribution in the domain of $[-1, 1]^{32}$. A schematic diagram of variable notations along with the network arrangement is given in Figure \ref{fig:fig_2}(right). While training the cyclic autoencoder, we utilize a sum of reconstruction loss on both $x_{real}$ and $z_{rand}$, i.e. $\mathcal{L}_{recon} = |x_{real}-x_{recon}| + |z_{rand} - z_{recon}|$. This reduces the reconstruction loss more aggressively to a lower value even with a limited 32-dimensional embedding space as compared to the counterpart with adversarial discriminator loss. This improves quality of pose generation but lacks in learning of a continuous embedding space, which can facilitate an efficient interpolation and traversal. Hence, in the second step of the learning protocol we introduce discriminator loss using the discriminator network, $Pose^{disc}$ with a combined objective function as, $ \mathcal{L} = \mathcal{L}_{recon} + \lambda\mathcal{L}_{adv} $. Here,

\begin{figure*}
\begin{center}
   \includegraphics[width=0.95\linewidth]{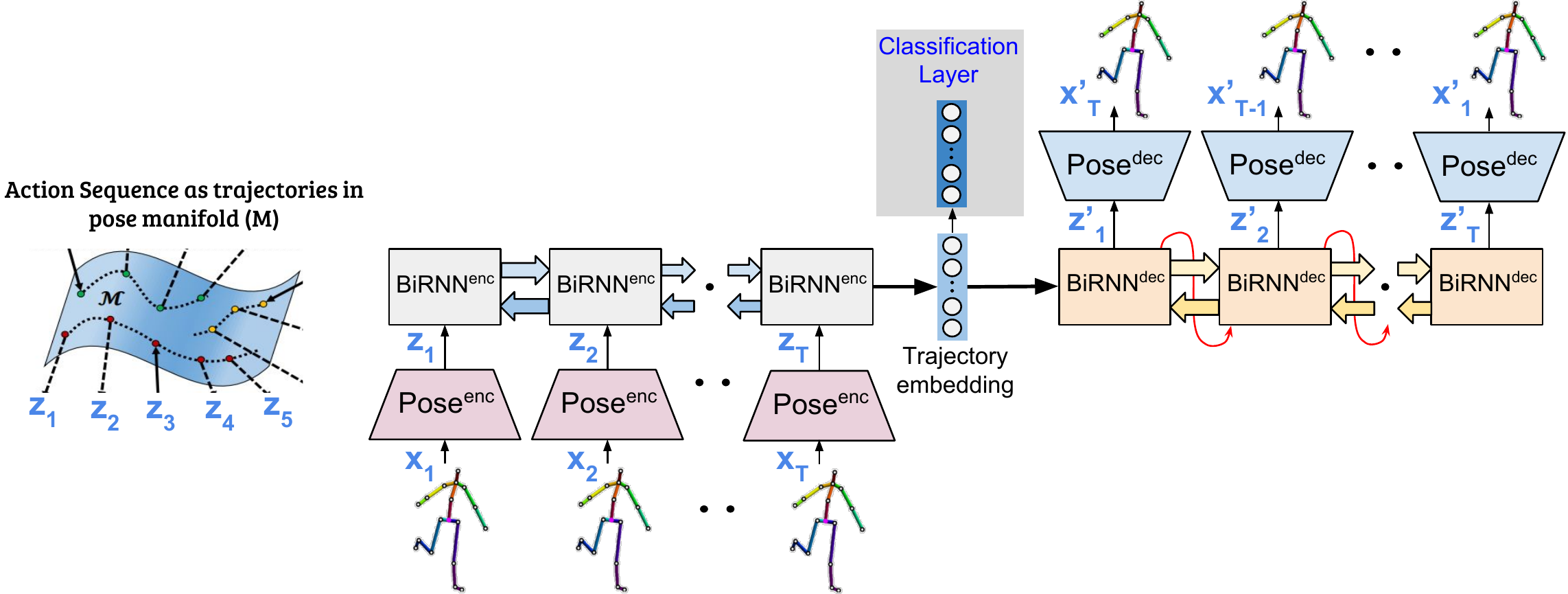}
 	\caption{End-to-end architecture of \textit{PoseRNN} for learning actions as a trajectory in pose manifold, enabling classification over the learned trajectory embedding}
 	\label{fig:fig_rnn}    
\end{center}
\vspace{-4mm}
\end{figure*}

\vspace{-4mm}
\begin{equation*}
\begin{split}
\mathcal{L}_{adv} =  & -\mathbb{E}_{X_{real}}[\log{(Pose^{disc}(X_{real}))}] \\
& -\mathbb{E}_{Z_{rand}}[\log{(1-(Pose^{disc}( Pose^{dec}(Z_{rand} ))))}]
\end{split}
\end{equation*}

However, we introduced $\mathcal{L}_{adv}$ initially with a very less weighting value of  $\lambda = \lambda_0$ as compared to the default $\mathcal{L}_{recon}$ loss. In further step, we gradually increase the value of weighting factor to $10*\lambda_0$. Note that, here $\mathcal{L}_{adv}$ is applied on the prediction, $X_{fake} = Pose^{dec}(Z_{rand} )$ against the true canonical pose distribution $p(x_{real})$.

\subsection{Learning trajectory embedding \textbf{\textit{PoseRNN}}}
\label{sec:posernn}
After learning the pose embedding manifold with both forward and backward projection networks (i.e. $Pose^{enc}$ and $Pose^{dec}$ respectively), we model pose dynamics as a trajectory in the embedding space as shown in the left section of Figure \ref{fig:fig_rnn}. An RNN encoder decoder architecture is incorporated to learn a representation which can embed the trajectory information in the previously learned pose manifold. Here, the final hidden representation of the encoder RNN is treated as a \textit{trajectory embedding}. Moreover, focusing on the final goal of action recognition to efficiently model both short-term and long-term pose dynamics, we employ a multi-layer bidirectional LSTM architecture~\cite{graves2005framewise} for both sequence encoder and decoder RNN i.e. $biRNN^{enc}$ and $biRNN^{dec}$ as shown in Figure \ref{fig:fig_rnn}. The concatenated output of both forward and backwards LSTM of first layer is fed as input sequence to the second layer bidirectional LSTM. The final trajectory embedding (or motion embedding) is obtained from the last layer as a function (fully-connected layer) of final hidden state representation of both forward and backward LSTM. Similarly for decoder RNN, we consider a bi-directional setup where initial hidden state of both forward and backwards LSTM is obtained as a function of the encoded trajectory embedding. Following Srivastava \etal~\cite{srivastava2015unsupervised}, for the decoder we consider chaining of previous prediction as input for the next time-step. Note that, while the forward LSTM decodes the trajectory in forwards direction, backward LSTM decodes it in reverse order. The final prediction of $z_{pose}$ sequence is obtained as a function of outputs from both forward and backward LSTM. 

While training \textit{PoseRNN}, considering the end goal of effective human motion prediction, we impose a direct loss on the predicted ${x}^\prime_{t}$ sequence i.e. ${x^\prime}_{t} = Pose^{dec}({z^\prime_t})$ with frozen parameters of $Pose^{dec}$ network from the \textit{EnGAN} training. Hence, instead of minimizing $\sum_t \vert {z^\prime}_t - z_t \vert$, we minimize the following loss function; $\mathcal{L}^{recon}_{RNN} = \sum_t \vert {x^\prime}_t - x_t \vert$. Additionally to enforce temporal consistency, we define $\delta x_t = \vert x_t - x_{t-1} \vert$ and similarly $\delta {x^\prime}_t$ is defined to formulate a first order smoothness loss as; $\mathcal{L}_{RNN}^{grad} = \sum_t \vert \delta {x^\prime}_t - \delta x_t \vert $. Finally the full \textit{PoseRNN} framework is trained using a combined loss function as, $\mathcal{L}_{RNN} = \mathcal{L}^{recon}_{RNN} + \hat{\lambda} \mathcal{L}_{RNN}^{grad}$. This greatly improves the motion reconstruction quality as compared to using only $\mathcal{L}^{recon}_{RNN}$ as the final loss function.

As the canonical pose representation does not contain any global view or translation information, the learned \textit{trajectory embedding} acquired from \textit{PoseRNN} is not enough to classify actions related to relative global position variations e.g. \textit{giving something to other person}, \textit{touch other person's pocket}, \textit{handshaking}, etc. Hence, global position and view information is provided as an additional input along with the output of $Pose^{enc}$ while training the full \textit{PoseRNN} pipeline. The corresponding reconstruction loss in also included in the updated final loss function, $\mathcal{L}_{RNN}$ with similar first order smoothness constraint.

To learn improved trajectory representation, which can encourage better action recognition results the \textit{PoseRNN} model should be able to capture local limb dynamics in a much efficient manner along with the global pose variations. Classes like \textit{eating}, \textit{drinking}, \textit{touching neck}, \textit{touching head}, etc. significantly involves local limb and specific skeleton joint dynamics in contrast to full-body pose variations inferred from the learned pose embedding. Furthermore, temporal interaction among the four limbs namely, two arms and two legs can also be leveraged explicitly to improve the modeling of local short-term trajectory dynamics. However recent challenging action recognition datasets also constitutes fine-grained categories related to very local joint dynamics such as \textit{typing}, \textit{writing}, \textit{playing with tablet}, etc. Therefore to address the above mentioned challenges we propose to use features from three different levels of hierarchy to efficiently model the local limb and joint level dynamics in the proposed \textit{PoseRNN} framework. As an input representation we fuse individual limb features acquired from the limb-level hierarchy of the learned $Pose^{enc}$ along with the global pose vector, $z$. Furthermore, the root-relative local joint coordinates are also included in the input representation to the final \textit{PoseRNN} framework. Thus, the $Pose^{enc}$ takes a concatenated feature representation consisting of the following; a) global pose embedding, b) four limb embedding features from $Pose^{enc}$, c) root-relative 3D joint coordinates after scale normalization and, d) parameters related to global positions (i.e translation information from raw hip coordinates and view information - sines and cosines - from Euler angles; $\alpha$, $\beta$ and $\gamma$). 

\section{Experimental evaluation}
\label{sec:experiment}
In this section we discuss about experimental evaluations to demonstrate  effectiveness of the proposed configuration of \textit{EnGAN}  and \textit{PoseRNN}, for 3D skeletal pose modelling and sequential trajectory learning respectively. We have trained the entire temporal pose modelling setup with enough data samples collected from various different datasets captured using Kinect device.

\subsection{Datasets and experimental settings}
\vspace{-0.5mm}
\noindent
\textbf{NTU RGB+D Dataset}~\cite{Shahroudy_2016_CVPR} This Kinect captured dataset is currently the largest dataset with RGB+D videos and skeleton data for human action recognition with 60 different action category annotations. The full dataset contains 56000 sample sequences across various diverse fine-grained categories related to daily activities also including different health-related actions. For each frame 25 joint coordinates is provided, which are captured from different camera views with a good diversity in face and body orientations. 
We have used the available Cross-View (CV) and Cross-Subject (CS) splits for fair evaluation of the proposed unsupervised feature learning framework against previous state-of-the-art methods. The given 60 action classes contains different fine actions related to local joint movements such as typing, writing etc. along with different multi person interaction based categories such as \textit{"giving something to other person"}, \textit{"punching other person"}, \textit{"walking towards a person"}, \textit{"pat on back of other person"}, etc. To demonstrate effectiveness of the learned trajectory embedding representation for the task of action recognition, we train a separate fully connected classification layer on the hidden temporal representations of $biRNN^{enc}$. 

\vspace{2mm}
\noindent
\textbf{SBU Kinect Interaction dataset}~\cite{yun2012two}: This Kinect captured dataset is an interaction dataset consisting of 282 sequence samples across 8 different classes. We use the standard subject independent 5-fold splits for evaluation of our unsupervised sequential pose modelling representation against previous state-of-the-art methods. To adapt the proposed \textit{PoseRNN} architecture for multi-person interaction, we first apply the former $biRNN^{enc}$  individually on both the sequence and then the latter classification layer is trained from the concatenated hidden layer activations acquired from the pose sequences of both the individuals. 

We train two different \textit{PoseRNN} frameworks for both 15 joint and 25 joint skeleton embeddings obtained from the corresponding \textit{EnGAN} training. For fair evaluation of transferability we used samples from PKU-MMD~\cite{liu2017pku} dataset for training both 15 joint and 25 joint \textit{EnGAN} setup. PKU-MMD dataset consists of 1076 video sequences across 51 action categories mostly in line with the categories of NTU RGB+D dataset. This ensures availability of enough diversity in individual pose samples and skeleton sequences for efficient modeling of local joints to global full body variations. 

\begin{table}[b]
\begin{center}
\vspace{-2mm}
\resizebox{\columnwidth}{!}{%
\begin{tabular}{|l|c|c|c|c|}
\hline
Training Scheme & $\mathcal{L}_{x_{recon}}$  & $\mathcal{L}_{z_{recon}}$ & $\mathcal{L}_{recon}$ & Acc \\
\hline\hline
GAN ($\mathcal{L}_{recon}$ + $\mathcal{L}_{adv}$) & 0.130 & \textbf{0.049} & 0.179 & 64.3\%\\
VAE ($\mathcal{L}_{x_{recon}}$ + $\mathcal{L}_{KLD}$) & 0.148 & 0.179 & 0.327 & 73.4\% \\
Auto Encoder & \textbf{0.051} & 0.245 & 0.297 & 87.3\% \\
Auto Encoder ($\mathcal{L}_{recon}$) + $\mathcal{L}_{adv}$ & 0.109 & 0.092  & 0.201 & 68.2\% \\
Proposed \textit{EnGAN} & 0.070 & 0.079  & \textbf{0.149} & \textbf{58.1\%} \\
\hline
\end{tabular}
}
\end{center}
\caption{Comparison of various training setups for \textit{EnGAN} on PKU-MMD Dataset over reconstruction capability and discriminability of Critic Network. (Lower is better) }
\vspace{-2mm}
\label{table:table_1}
\end{table}

\subsection{Ablation study}
\label{subsec:ablation}
\noindent
\textbf{Effectiveness of  \textit{EnGAN} framework:} We have performed experiments with different autoencoder setups. We also train a critic network, different from the discriminator network, to distinguish the samples generated by these setups, from the real ones. To measure the discriminability of the critic network for a given setup, we evaluate the accuracy of critic network on the task of classifying the generated samples as fake ones.
Note that here pose-samples are generated by sampling a random $z \sim P(z)$ (In this case, $U(-1, 1)$) and project it to x = $Pose^{enc}(z)$. Table \ref{table:table_1} shows the reconstruction errors and discriminability on a held-out set of diverse sample pose frames for different training protocols.

\begin{table}[b]
\vspace{-1mm}
\begin{center}
\begin{tabular}{|l|c|c|}
\hline
Input Features & NTU & PKU \\ 
\hline\hline
$(P_{J})$ &  66.1 \% & 75.0 \% \\
$(E_{P})$ &  71.3 \% & 77.3 \% \\
$(P_{J})$ + $(E_{P})$ &  74.8 \% & 82.4\% \\
\textbf{$(P_{J})$ + $(E_{P})$ + $(E_{L})$} &  \textbf{78.7} \% & \textbf{85.9} \% \\
\hline
\end{tabular}
\end{center}
\caption{Comparison of feature fusion techniques.
$(P_{J})$-Joint Positions 
$(E_{P})$-Pose Embeddings 
$(E_{L})$-Limb Embeddings
}
  \label{table:table_2}
\end{table}

\begin{figure*}[!t]
\begin{center}
	\includegraphics[width=0.96\textwidth]{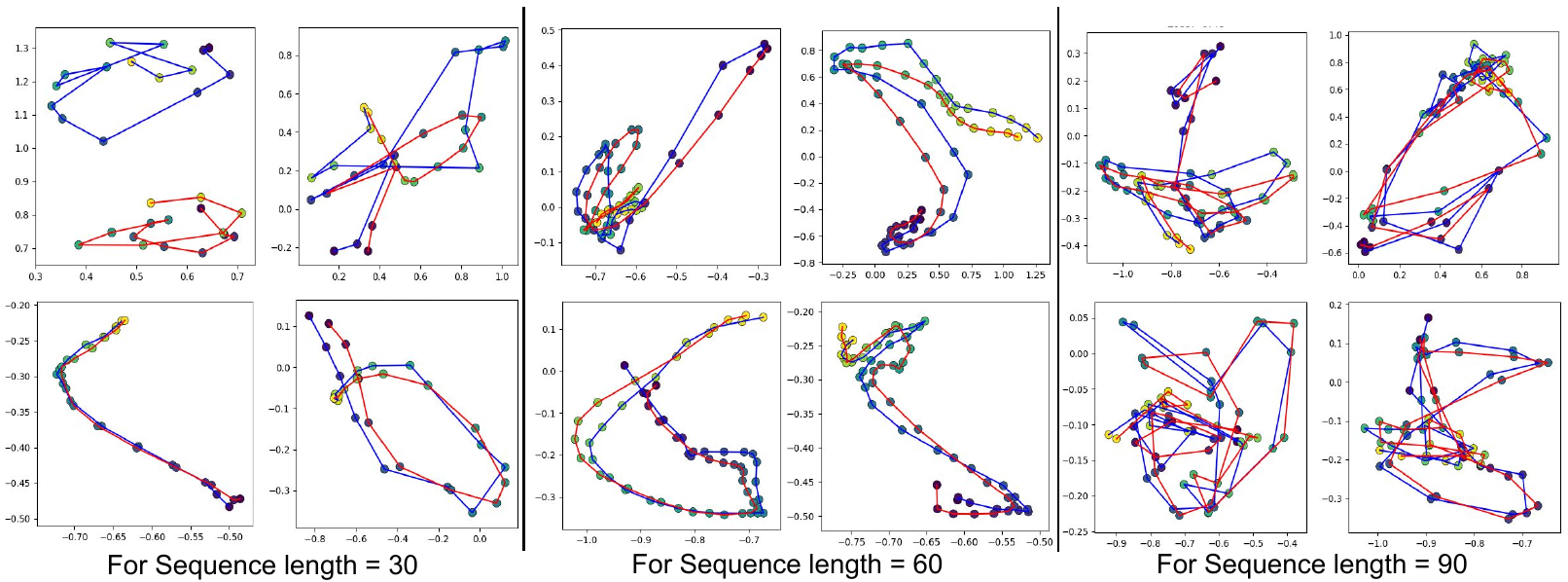}
 	\caption{Illustrations of the pose manifold trajectories of action sequences - Ground Truth (Blue) and Reconstructed (Red) by \textit{PoseRNN}, projected to 2D via PCA. The top 2 illustrations for sequence length 30, represents trajectories with highest reconstruction losses.}
 	\label{fig:fig_trajectory}   
 	\vspace{-5mm}
\end{center}
\end{figure*}

\begin{figure*}[t]
\begin{center}
	\includegraphics[width=1.0\textwidth]{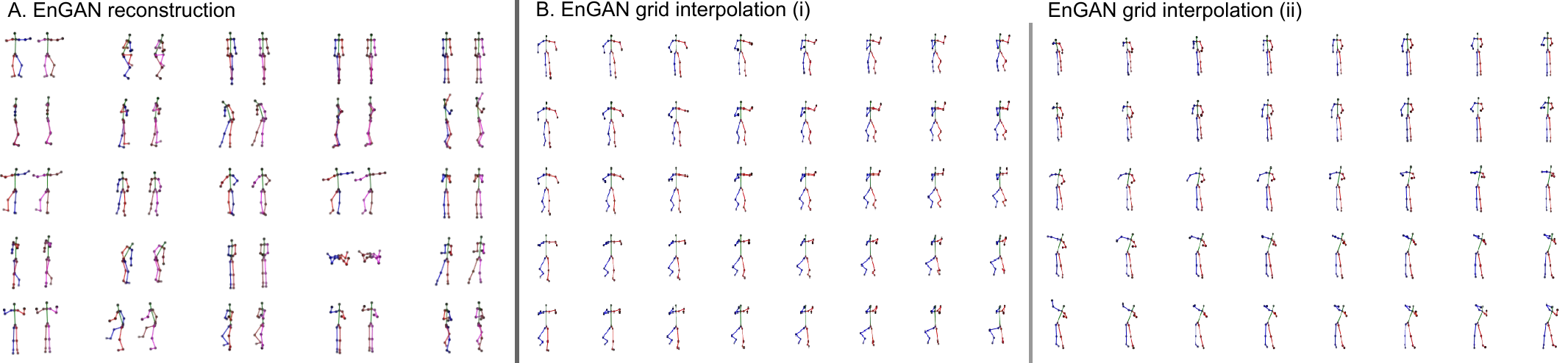}
 	\caption{Illustrations of the pose reconstructions quality (left) and Grid Interpolation of Skeleton Poses (right) generated from a continuous pose-embedding manifold, by \textit{EnGAN}. Note that, in the reconstruction results on the left side of the figure, each pair of skeleton represents the ground truth skeleton (left) and reconstructed skeleton (right), respectively.}
 	\label{fig:fig_quality}   
 	\vspace{-5mm}
\end{center}
\end{figure*}

First we trained the complete \textit{EnGAN} setup with both $\mathcal{L}_{recon}$ and $\mathcal{L}_{adv}$ from scratch. As it is clear from the metrics in Table \ref{table:table_1} that although the first setup achieves better performance as compared to a simple autoencoder with only $\mathcal{L}_{recon}$ loss, it performs quite sub-optimally for skeletal reconstruction, as measured by  $\mathcal{L}_{x_{recon}} = |x_{real}-x_{recon}|$. In contrast to that, a plain autoencoder setup, even though performing extremely well on just skeleton reconstruction, shows expectedly suboptimal performance on learning a continuous pose-manifold, as measured by $\mathcal{L}_{z_{recon}} = |z_{rand} - z_{recon}|$ and critic accuracy. We report results on the proposed training scheme (\textit{EnGAN}) of gradually increasing the weighting factor $\lambda$ associated with $\mathcal{L}_{adv}$ in the combined loss function $\mathcal{L}$. The proposed training protocol also improves reconstruction loss (Figure \ref{fig:fig_quality} a) even better than the autoencoder setup by enabling learning of a continuous pose-manifold (Figure \ref{fig:fig_quality} b) along with an efficient one-shot transformation from skeletal-pose space $X$ to the latent space $Z$.

\vspace{2mm}
\noindent
\textbf{Effectiveness of trajectory embedding learned using \textit{PoseRNN}:}
We evaluated multiple \textit{PoseRNN} setups to obtain the best trajectory embedding representation, which can reproduce the pose sequence with minimum reconstruction loss. Effectiveness of different input representations ($P_{J}$: Joint Positions, $E_{P}$: Pose Embeddings, $E_{L}$: Limb Embeddings) to the proposed \textit{PoseRNN} setup is demonstrated in Table \ref{table:table_2}. Use of only joint coordinates $(P_{J})$ without incorporating \textit{EnGAN} into the \textit{PoseRNN} framework delivers suboptimal performance. This validates the effectiveness of disentangled learning of pose and trajectory embedding for both pose synthesis and unsupervised feature learning task. 

It is also observed that due to the disentanglement of the pose learning from sequence learning task, reconstructions of the pose embedding sequences by the proposed \textit{PoseRNN} network scales to number of frames as high as 120, which is highly unlikely for end-to-end reconstruction framework ~\cite{li2017auto,taku_motion_synthesis} (show comparatively higher losses for reconstruction).
Pose manifold trajectories of action sequences (blue - ground truth, red - reconstructed) are illustrated in Figure \ref{fig:fig_trajectory}. It is observed that the model is able to reconstruct and generate pose trajectories fairly well for sequences as long as 120 frames, and hence efficiently captures inherent features of an action sequence. 

\begin{table}[b]\small
\vspace{-2mm}
\begin{center}
\begin{tabular}{|l|c|c|}
\hline
Methods & Average $L_{x_{recon}}(t)$  \\ 
\hline\hline
\textit{PoseRNN}(baseline) & 0.458  \\
{Holden~\etal~\cite{holden2015learning}} & 0.402  \\
\textit{EnGAN-PoseRNN} & \textbf{0.342}   \\

\hline
\end{tabular}
\end{center}
\caption{Comparisons of time average reconstruction loss on the prediction of final skeleton pose on NTU dataset (120 frames).}
  \label{table:table_5a}
\end{table}

\subsection{Comparison with existing approaches}
\label{subsec:ablation}

\noindent
\textbf{Effectiveness of \textit{ENGAN-PoseRNN}:}
The proposed motion modeling framework \textit{EnGAN-PoseRNN} is compared against a baseline, \textit{PoseRNN}(baseline) taking raw canonical skeleton joint coordinates by avoiding learning of pose embedding (\textit{ENGAN}). We also include comparison against the convolutional auto-encoder proposed by Holden \etal~\cite{holden2015learning}, by training it on the extracted canonical skeleton joint locations from NTU RGB+D dataset with input sequence length and dimensions similar to \textit{EnGAN-PoseRNN} setup. We follow the exact architecture, and regularized loss function as proposed in ~\cite{holden2015learning}. Table \ref{table:table_5a} holds the average reconstruction loss of predicted poses over 120 frames on the test set. It clearly demonstrates superior expressibility of the proposed \textit{EnGAN-PoseRNN}  model in encoding the entire motion sequence in a single trajectory embedding vector.

\begin{table}[b!]\small
\begin{center}
\vspace{-2mm}
\begin{tabular}{|l|c|c|c|}
\hline
Methods  & \begin{tabular}[c]{@{}c@{}}Feature \\ supervision\end{tabular} & CS & CV \\ 
\hline\hline
ST-LSTM~\cite{liu2016spatio} &  Full & 69.2 &  77.7 \\
STA-LSTM~\cite{song2017end} &  Full & 73.4  &  81.2  \\
TS-LSTM~\cite{Lee_2017_ICCV} &  Full & 75.9 &  82.5 \\

GCA-LSTM~\cite{liu2017global} & Full & 74.4  & 82.8  \\
URNN-2L-T~\cite{li2017adaptive} & Full & 74.6  & 83.2  \\
TPNet~\cite{hu2017temporal} & Full & 75.3 &  84.0  \\
VA-LSTM~\cite{zhang2017view} & Full & \textbf{79.4} & \textbf{87.6} \\
\hline \hline

\textit{VAE-PoseRNN} &  Unsup. & 56.4 &  63.8  \\
{}  &  Semi & 61.2  &  69.8 \\ \hline

\textit{PoseRNN}(baseline) &  Unsup. & 59.8 &  69.0  \\
{}  &  Semi & 69.7  &  77.9 \\ \hline

{Holden~\etal~\cite{holden2015learning}} &  Unsup. & 61.2 &  70.2  \\
{} &  Semi & 72.9  &  81.1 \\ \hline

\textit{EnGAN-PoseRNN} &  Unsup. & 68.6 &  77.8  \\
{} &  Semi & \textbf{78.7}  &  \textbf{86.5} \\

\hline
\end{tabular}
\end{center}
\caption{Comparisons on the NTU dataset, given  feature supervision level, for standard Cross-Subject and Cross-View settings}
\vspace{-3mm}
  \label{table:table_5}
\end{table}

\vspace{2mm}
\noindent
\textbf{Effectiveness of motion embedding for action recognition on NTU:}
Following the literature of unsupervised feature learning ~\cite{pathakCVPR17learning,pathakCVPR16context}, we compose two different settings viz. a) Unsupervised and b) Semi-supervised.

In unsupervised setting, a single fully-connected classification layer (considering linear classifier) is trained on the final trajectory embedding vector, i.e. the output of $biRNN^{enc}$. To effectively handle interaction based action categories, the classification layer is trained on the concatenated trajectory embedding from two different motion sequences. For single subject based action categories, we repeat the same motion sequence two times before feeding it to the final classification layer. Moreover, we train a single classification layer for both interaction and non-interaction based action categories. We use only 40\% of the labelled training samples as it is enough to train a single fully-connected network as compared to the full 100\% used by other fully supervised methods. Note that, while training in this setting, the parameters of the underlying $Pose^{enc}$ followed by $biRNN^{enc}$ is kept frozen to evaluate discriminability of the unsupervisely learned motion embedding.

In semi-supervised setting, we allow parameters of $biRNN^{enc}$ (i.e. fine-tuning) to update along with the previously introduced classification layer (initialized from unsupervised setting) on 40\% of the labelled training samples. Here, semi-supervised refers to use of semi-supervisedly learned motion embedding in contrast to the previous unsupervised setting.

Table \ref{table:table_5} holds comparison of \textit{EnGAN-PoseRNN} on both unsupervised and semi-supervised setting against the same settings of the convolutional autoencoder proposed by Holden \etal~\cite{holden2015learning}. Results clearly highlight superiority of the proposed \textit{EnGAN-PoseRNN} against the unsupervised framework proposed by Holden \etal. We also report accuracies of other fully supervised approaches on both cross-subject (CS) and cross-view (CV) setup to demonstrate competitive state-of-the-art performance.


\vspace{2mm}
\noindent
\textbf{Effectiveness of transferability on SBU}
Following the motivation regarding the need of an unsupervised feature learning framework, we plan to demonstrate transferability (or generalixzability) of the learned trajectory embedding feature from NTU to SBU dataset. Instead of training \textit{PoseRNN} on SBU dataset, we plan to evaluate discriminability of the learned embedding trained from NTU dataset. For fair comparison, we first train \textit{TS-LSTM}~\cite{Lee_2017_ICCV} and \textit{ST-LSTM}~\cite{liu2016spatio} (using the publicly available code) with full supervision on NTU dataset. Then we discard the final NTU classification layer and trained a newly introduced SBU classification layer on 40\% of labelled SBU training set. Following the previous unsupervised and semi-supervised setting, we introduced two different settings named a) unsupervised transfer, and b) semi-supervised transfer. In unsupervised transfer, only the last classification layer is trained using 40\% of the labelled SBU train set. Whereas, in semi-supervised transfer, the full network is fine-tuned after initialization from the unsupervised transfer framework. 

Table \ref{table:table_6} shows classification accuracy on unsupervised and semi-supervised transfer setting, for both supervisedly learned motion feature (\textit{TS-LSTM} and \textit{ST-LSTM}) and unsupervisedly learned motion feature (Holden \etal and \textit{EnGAN-PoseRNN}). This clearly demonstrates generalizability of learned embedding from the proposed unsupervised learning framework.


\begin{table}[t]\small
\begin{center}
\begin{tabular}{|l|c|c|c|c|}
\hline
Methods & \begin{tabular}[c]{@{}c@{}}Unsupervised \\ transfer\end{tabular} & \begin{tabular}[c]{@{}c@{}}Semi-supervised \\ transfer\end{tabular} \\ 
\hline\hline
ST-LSTM~\cite{liu2016spatio}  & 73.1 &  87.1 \\
TS-LSTM~\cite{Lee_2017_ICCV}  & 72.7 &  86.5 \\

\hline\hline
\textit{PoseRNN}(baseline) & 73.1 &  81.8  \\
{Holden~\etal~\cite{holden2015learning}} & 73.4 & 82.6 \\
\textit{EnGAN-PoseRNN} & \textbf{77.0} & \textbf{88.2}  \\

\hline
\end{tabular}
\end{center}
\caption{Comparison of transfer ability (NTU to SBU) of unsupervisedly learned motion feature against the supervised counterpart. Note that fine-tuning in semi-supervised setting is performed according to the performance on a held-out validation set.}
\vspace{-3mm}
  \label{table:table_6}
\end{table}




\section{Conclusion}
In this paper, we have proposed a generative model for unsupervised feature representation learning of 3D Human Skeleton poses and action sequences. Experimental results and visualizations show qualitative strengths of the proposed framework for temporal pose generation. We demonstrate effectiveness and transferability of the learned trajectory embedding by empirical evaluation for the same task of fine-grained action recognition on multiple datasets, considering interaction based categories as well. Performance on standard action recognition datasets for both unsupervised and semi-supervised setup are competitive with fully supervised state-of-the-art methods. 

\vspace{2mm}
\noindent
\textbf{Acknowledgements} This work was supported by a CSIR Fellowship (Jogendra), and a project grant from Robert Bosch Centre for Cyber-Physical Systems, IISc.

{\small
\bibliographystyle{ieee}
\bibliography{ms}

\begin{thebibliography}{10}\itemsep=-1pt

\bibitem{akhter2015pose}
I.~Akhter and M.~J. Black.
\newblock Pose-conditioned joint angle limits for 3d human pose reconstruction.
\newblock In {\em Proceedings of the IEEE Conference on Computer Vision and
  Pattern Recognition}, pages 1446--1455, 2015.

\bibitem{butepage2017deep}
J.~B{\"u}tepage, M.~J. Black, D.~Kragic, and H.~Kjellstr{\"o}m.
\newblock Deep representation learning for human motion prediction and
  classification.
\newblock In {\em IEEE Conference on Computer Vision and Pattern Recognition
  (CVPR)}, page 2017, 2017.

\bibitem{liu2017pku}
L.~Chunhui, H.~Yueyu, L.~Yanghao, S.~Sijie, and L.~Jiaying.
\newblock Pku-mmd: A large scale benchmark for continuous multi-modal human
  action understanding.
\newblock {\em ACM Multimedia workshop}, 2017.

\bibitem{du2016representation}
Y.~Du, Y.~Fu, and L.~Wang.
\newblock Representation learning of temporal dynamics for skeleton-based
  action recognition.
\newblock {\em IEEE Transactions on Image Processing}, 25(7):3010--3022, 2016.

\bibitem{du2015hierarchical}
Y.~Du, W.~Wang, and L.~Wang.
\newblock Hierarchical recurrent neural network for skeleton based action
  recognition.
\newblock In {\em Proceedings of the IEEE conference on computer vision and
  pattern recognition}, pages 1110--1118, 2015.

\bibitem{fragkiadaki2015recurrent}
K.~Fragkiadaki, S.~Levine, P.~Felsen, and J.~Malik.
\newblock Recurrent network models for human dynamics.
\newblock In {\em Computer Vision (ICCV), 2015 IEEE International Conference
  on}, pages 4346--4354. IEEE, 2015.

\bibitem{graves2005framewise}
A.~Graves and J.~Schmidhuber.
\newblock Framewise phoneme classification with bidirectional lstm and other
  neural network architectures.
\newblock {\em Neural Networks}, 18(5-6):602--610, 2005.

\bibitem{holden2015learning}
D.~Holden, J.~Saito, T.~Komura, and T.~Joyce.
\newblock Learning motion manifolds with convolutional autoencoders.
\newblock In {\em SIGGRAPH Asia 2015 Technical Briefs}, page~18. ACM, 2015.

\bibitem{hu2017temporal}
Y.~Hu, C.~Liu, Y.~Li, S.~Song, and J.~Liu.
\newblock Temporal perceptive network for skeleton-based action recognition.
\newblock In {\em Proc. Brit. Mach. Vis. Conf}, pages 1--2, 2017.

\bibitem{taku_motion_synthesis}
T.~Komura, I.~Habibie, D.~Holden, J.~Schwarz, and J.~Yearsley.
\newblock {\em A Recurrent Variational Autoencoder for Human Motion Synthesis}.
\newblock 7 2017.

\bibitem{Lee_2017_ICCV}
I.~Lee, D.~Kim, S.~Kang, and S.~Lee.
\newblock Ensemble deep learning for skeleton-based action recognition using
  temporal sliding lstm networks.
\newblock In {\em The IEEE International Conference on Computer Vision (ICCV)},
  Oct 2017.

\bibitem{li2017adaptive}
W.~Li, L.~Wen, M.-C. Chang, S.~N. Lim, and S.~Lyu.
\newblock Adaptive rnn tree for large-scale human action recognition.

\bibitem{li2017auto}
Z.~Li, Y.~Zhou, S.~Xiao, C.~He, and H.~Li.
\newblock Auto-conditioned lstm network for extended complex human motion
  synthesis.
\newblock {\em arXiv preprint arXiv:1707.05363}, 2017.

\bibitem{liu2016spatio}
J.~Liu, A.~Shahroudy, D.~Xu, and G.~Wang.
\newblock Spatio-temporal lstm with trust gates for 3d human action
  recognition.
\newblock In {\em European Conference on Computer Vision}, pages 816--833.
  Springer, 2016.

\bibitem{liu2017global}
J.~Liu, G.~Wang, P.~Hu, L.-Y. Duan, and A.~C. Kot.
\newblock Global context-aware attention lstm networks for 3d action
  recognition.
\newblock In {\em CVPR}, 2017.

\bibitem{martinez2017human}
J.~Martinez, M.~J. Black, and J.~Romero.
\newblock On human motion prediction using recurrent neural networks.
\newblock In {\em 2017 IEEE Conference on Computer Vision and Pattern
  Recognition (CVPR)}, pages 4674--4683. IEEE, 2017.

\bibitem{pathakCVPR17learning}
D.~Pathak, R.~Girshick, P.~Doll{\'a}r, T.~Darrell, and B.~Hariharan.
\newblock Learning features by watching objects move.
\newblock In {\em CVPR}, 2017.

\bibitem{pathakCVPR16context}
D.~Pathak, P.~Kr\"ahenb\"uhl, J.~Donahue, T.~Darrell, and A.~Efros.
\newblock Context encoders: Feature learning by inpainting.
\newblock 2016.

\bibitem{shahroudy2016ntu}
A.~Shahroudy, J.~Liu, T.-T. Ng, and G.~Wang.
\newblock Ntu rgb+ d: A large scale dataset for 3d human activity analysis.
\newblock {\em arXiv preprint arXiv:1604.02808}, 2016.

\bibitem{Shahroudy_2016_CVPR}
A.~Shahroudy, J.~Liu, T.-T. Ng, and G.~Wang.
\newblock Ntu rgb+d: A large scale dataset for 3d human activity analysis.
\newblock In {\em The IEEE Conference on Computer Vision and Pattern
  Recognition (CVPR)}, June 2016.

\bibitem{song2017end}
S.~Song, C.~Lan, J.~Xing, W.~Zeng, and J.~Liu.
\newblock An end-to-end spatio-temporal attention model for human action
  recognition from skeleton data.
\newblock In {\em AAAI}, volume~1, page~7, 2017.

\bibitem{srivastava2015unsupervised}
N.~Srivastava, E.~Mansimov, and R.~Salakhudinov.
\newblock Unsupervised learning of video representations using lstms.
\newblock In {\em International conference on machine learning}, pages
  843--852, 2015.

\bibitem{wang2017modeling}
H.~Wang and L.~Wang.
\newblock Modeling temporal dynamics and spatial configurations of actions
  using two-stream recurrent neural networks.
\newblock In {\em e Conference on Computer Vision and Pa ern Recognition
  (CVPR)}, 2017.

\bibitem{yeh2016semantic}
R.~Yeh, C.~Chen, T.~Y. Lim, M.~Hasegawa-Johnson, and M.~N. Do.
\newblock Semantic image inpainting with perceptual and contextual losses.
\newblock {\em arXiv preprint arXiv:1607.07539}, 2016.

\bibitem{yun2012two}
K.~Yun, J.~Honorio, D.~Chattopadhyay, T.~L. Berg, and D.~Samaras.
\newblock Two-person interaction detection using body-pose features and
  multiple instance learning.
\newblock In {\em Computer Vision and Pattern Recognition Workshops (CVPRW),
  2012 IEEE Computer Society Conference on}, pages 28--35. IEEE, 2012.

\bibitem{zhang2017view}
P.~Zhang, C.~Lan, J.~Xing, W.~Zeng, J.~Xue, and N.~Zheng.
\newblock View adaptive recurrent neural networks for high performance human
  action recognition from skeleton data.
\newblock In {\em IEEE International Conference on Computer Vision}, 2017.

\bibitem{zhu2017unpaired}
J.-Y. Zhu, T.~Park, P.~Isola, and A.~A. Efros.
\newblock Unpaired image-to-image translation using cycle-consistent
  adversarial networks.
\newblock In {\em IEEE International Conference on Computer Vision}, 2017.

\bibitem{zhu2016co}
W.~Zhu, C.~Lan, J.~Xing, W.~Zeng, Y.~Li, L.~Shen, X.~Xie, et~al.
\newblock Co-occurrence feature learning for skeleton based action recognition
  using regularized deep lstm networks.
\newblock In {\em AAAI}, volume~2, page~8, 2016.

\end{thebibliography}
}

\end{document}